\documentclass[sigconf,authorversion]{acmart}
\renewcommand\footnotetextcopyrightpermission[1]{} % removes footnote with conference information in first column
\usepackage{graphicx}
\usepackage{amsmath}
\usepackage{booktabs}
\usepackage{colortbl}
\usepackage[linesnumbered,ruled,vlined]{algorithm2e}
\usepackage{hyperref}
\usepackage{url}
\usepackage[utf8]{inputenc} % allow utf-8 input
\usepackage[T1]{fontenc}    % use 8-bit T1 fonts
\usepackage{amsfonts}       % blackboard math symbols
\usepackage{nicefrac}       % compact symbols for 1/2, etc.
\usepackage{microtype}      % microtypography
\usepackage{xcolor}         % colors
\usepackage{wrapfig}
\usepackage{multirow}
\usepackage{dsfont}
\usepackage{setspace}
\usepackage{caption}
\usepackage{subfigure}
\usepackage{fontawesome}
\usepackage{bbm}
\AtBeginDocument{%
  \providecommand\BibTeX{{%
    \normalfont B\kern-0.5em{\scshape i\kern-0.25em b}\kern-0.8em\TeX}}}

\newcommand*{\method}{CORE}

\begin{document}

% \title{U-CORE: A Universal COhort-based Representation Enhancement Framework for EHR Representation Learning}
\title{Towards Cohort Intelligence: A Universal Cohort Representation \\ Learning Framework for Electronic Health Record Analysis}

% A Universal Cohort EHR Representation Learning Framework for Healthcare Analytics

%    <concept>
%        <concept_id>10010147.10010257.10010293.10010319</concept_id>
%        <concept_desc>Computing methodologies~Learning latent representations</concept_desc>
%        <concept_significance>500</concept_significance>
%        </concept>
%  </ccs2012>

% \ccsdesc[500]{Computing methodologies~Learning latent representations}

%    <concept>
%        <concept_id>10010147.10010257.10010293.10010319</concept_id>
%        <concept_desc>Computing methodologies~Learning latent representations</concept_desc>
%        <concept_significance>500</concept_significance>
%        </concept>
%    <concept>
%        <concept_id>10010405.10010444.10010449</concept_id>
%        <concept_desc>Applied computing~Health informatics</concept_desc>
%        <concept_significance>500</concept_significance>
%        </concept>
%  </ccs2012>

% \ccsdesc[500]{Computing methodologies~Learning latent representations}
% \ccsdesc[500]{Applied computing~Health informatics}

% \ccsdesc[500]{General and reference~Design}
% \ccsdesc[300]{Computing methodologies~Semantic networks}
% \ccsdesc[300]{Computing methodologies~Unsupervised learning}

\author{Changshuo Liu}
\affiliation{%
  \institution{National University of Singapore}
  \country{Singapore}}
\email{e0792475@u.nus.edu}

\author{Wenqiao Zhang}
\affiliation{%
  \institution{National University of Singapore}
  \country{Singapore}}
\email{wenqiao@nus.edu.sg}

\author{Beng Chin Ooi}
\affiliation{%
  \institution{National University of Singapore}
  \country{Singapore}}
\email{ooibc@comp.nus.edu.sg}

\author{James Wei Luen Yip}
\affiliation{%
  \institution{National University Health System}
  \country{Singapore}}
\email{james_yip@nuhs.edu.sg}

\author{Lingze Zeng}
\affiliation{%
  \institution{National University of Singapore}
  \country{Singapore}}
\email{lingze@nus.edu.sg}

\author{Kaiping Zheng}
\affiliation{%
  \institution{National University of Singapore}
  \country{Singapore}}
\email{kaiping@comp.nus.edu.sg}

\keywords{cohort, representation learning, healthcare, universal framework}


% kp: for discussion in person
% 1. maybe emphasize “backbone” is a model; so CORE can be generally applicable to different backbones
% 2. what is the patient similarity task? why is it semi-supervised?
% 3. usage of “framework” – “module”: CORE is a framework, contains 2 modules; but it is stated that CORE can be integrated into backbones as a module
% 4. RW: diff vs GRASP – should be rephrased: e.g., a major difference is CORE models intra-cohort info
% 5. notations: quite complex, can skip “p” and “v” sometimes for simplicity – better keep within 2 levels
% 6. why cohort modeling using visit-level representations, rather than patient-level representations?
% 7. sec3.4.1 should end with Intra() formula
% 8. sec3.4.2 should include Inter() formula
% 9. inputs: notes or codes?
% 10. dataset usage - why select UCI one?

\begin{abstract}
  Electronic Health Records (EHR) are generated from clinical routine care recording valuable information of broad patient populations, which provide plentiful opportunities  for improving patient management and intervention strategies in clinical practice.
  To exploit the enormous potential of EHR data, a popular EHR data analysis paradigm in machine learning is EHR representation learning, which first leverages the individual patient's EHR data to learn informative representations by a backbone, and supports diverse health-care downstream tasks grounded on the representations.
  Unfortunately, such a paradigm fails to access the in-depth analysis of  patients' relevance, which is generally known as cohort studies in clinical practice.
  % kp: if patient-level cohort, then remove "or visits"
  % kp: "approach" or "paradigm"? may need to be consistent
  Specifically, patients in the same cohort tend to share similar characteristics, implying
  %they have a similar trend in healthcare, such as similar symptoms or diseases.
  their resemblance in medical conditions such as symptoms or diseases.
  % the shared characteristics across patients, the potential and valuable,  which limit EHR data utilization.
  % similar patients, also known as cohorts in healthcare, which leads to limited performance.
  % Therefore, it's of great value to explore insights into EHR data. With the rapid development of machine learning, there are a great many methods or models proposed to obtain the representations of patients from EHR data. 
  % But previous methods ignore the priceless information of similar patients, also known as cohorts in healthcare, which leads to limited performance.
  In this paper, we 
  % address such an overlooked perspective in EHR representation learning, and correspondingly 
  propose a universal \textbf{CO}hort \textbf{R}epresentation l\textbf{E}arning (CORE) framework to augment EHR utilization by leveraging the fine-grained cohort information among patients. 
  % kp: "to refine the previous coarse-grained individual EHR utilization." -> need 1 sentence to summarize the limitations of existing work - as this is not the 1st work using cohort info
  %% wq: refined
  % \method{} aims to extract cohorts that rely on a pre-context task and encode the information of cohorts into backbone models.
  % To address the problem, we propose \method{}, a \textbf{U}niversal \textbf{CO}hort-based \textbf{R}epresentation \textbf{E}nhancement Framework for EHR Representation Learning.
  In particular, \method{} first develops an explicit patient modeling task 
  %%according to
  based on the prior knowledge of patients' diagnosis codes, which measures the latent relevance among patients to adaptively divide the cohorts for each patient. Based on the constructed cohorts, \method{} recodes the pre-extracted EHR data representation from intra- and inter-cohort perspectives, yielding augmented EHR data representation learning.
  % kp: the subject better be consistent, either "CORE" or "we"
  \method{} is readily applicable to diverse backbone models, serving as a universal plug-in framework to infuse cohort information into healthcare methods for boosted performance.
  % kp: pls check above
  % We prove that the novel patient cohorts indeed enhance EHR representations.
  % Experimental results with two medical datasets on readmission tasks demonstrate \method{} outperforms GRASP and other baselines. 
  We conduct an extensive experimental evaluation on two real-world datasets, and the experimental results demonstrate the effectiveness and generalizability of \method{}.
\end{abstract}
\maketitle
\section{Introduction}
Electronic health records (EHR) data are electronically stored medical histories of patients in a healthcare system. EHR data typically consists of patients' demographics (\emph{e.g.}, age, gender), and temporal medical variables (\emph{e.g.}, diagnoses, medications, laboratory tests, and clinical notes)~\cite{johnson2016mimic}.
Therefore, EHR data are beneficial to power data-driven approaches in the healthcare community and facilitate critical clinical decision-making (\emph{e.g.}, mortality
prediction, patient subtyping, and diagnosis prediction) for more optimized patient management~\cite{yan2020interpretable, hou2020predicting, baytas2017patient, wynants2020prediction}, including assisting doctors in analyzing patients' health conditions, developing treatment plans, and preventing adverse outcomes in a more intelligent and effective way, among others.

% Electronic Health Records (EHR) data are the digital version of patients' medical histories in a healthcare system, 
% Each patient may have several admissions to the hospital in EHR data.
% EHR data are the main data source for health analysis and disease prediction. 
% In general, EHR data are temporally represented by a set of medical variables, which describes patients' detailed information, such as age, gender, diagnosis, medication, laboratory tests, clinical notes, and medical images captured from kinds of medical machines. 

% Therefore, fully utilizing the valuable EHR data for healthcare tasks is of vital importance.
% One cohort contains patients whose ages are between 60 and 90. The other cohort contains patients who had a fever.
\begin{figure*}[t]
  \centering
  \includegraphics[width=\linewidth]{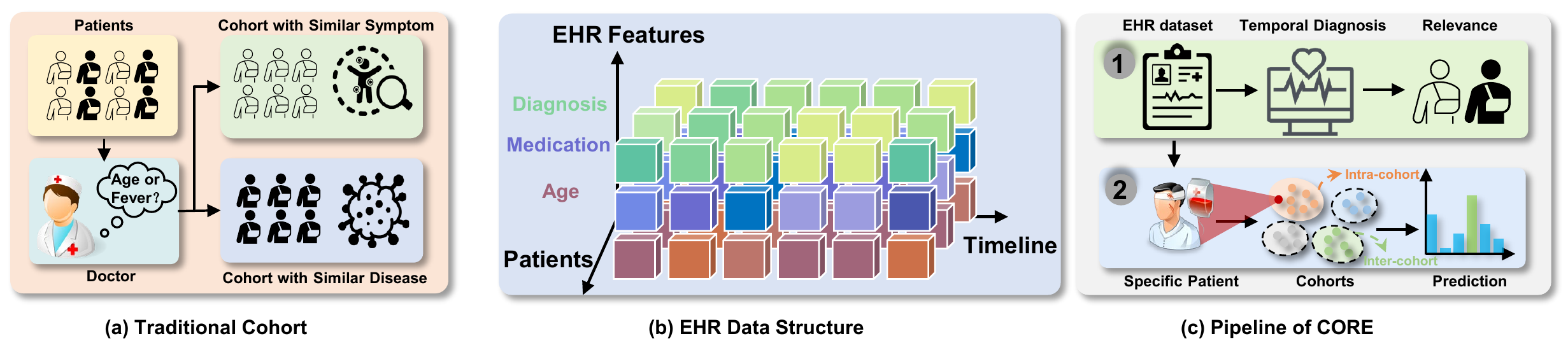}
  \caption{ (a) describes traditional cohort study in healthcare. (b) describes the complex EHR data structure from three dimensions. (c) illustrates the two steps of \method{}.  }
  % kp: current caption is only for (a)? in (b): "Age" -> "Demographics"? in (c): add Step1 and Step2 in figure?
  %% wq: revised
  \label{fig:cohort_intro}
\end{figure*}

{
Contemporary approaches~\cite{huang2019clinicalbert, choi2017gram, shang2019pre} generally learn the generalizable representation by exploiting the potential information of EHR data and then using these learned representations to facilitate different downstream healthcare tasks. Unfortunately, one indispensable consideration to achieve effective EHR representation learning has been overlooked by most existing studies - \textbf{Patient Cohorts}. 
%In healthcare, a patient cohort is a term used in medical research to define groupings of individuals with common traits, such as age and health factors.
% kp: "healthcare" vs "medical research"? changed to below
% wq: changed
In healthcare research, a patient cohort defines groupings of individuals with common traits, such as demographics, and health conditions.
For example, as shown in Figure~\ref{fig:cohort_intro}, when COVID-19 ravages the world, if doctors select COVID-19 patients whose ages are between 60 and 90 to derive the cohort, doctors may find that these 
patients tend to exhibit similarly severe symptoms. In addition, if doctors select the cohort according to the symptoms of the patients, they may find the patients who have a fever tend to develop a similar disease, \emph{i.e.}, COVID-19. In practice, patient cohorts are integral to researching and developing effective medical interventions.
Based on the aforementioned insights, a meaningful optimization goal of EHR representation learning is to explore an effective cohort construction approach that evolves from \emph{individual intelligence} to \emph{cohort intelligence}, thereby unleashing the potential of EHR data profoundly to facilitate EHR analysis further.

As illustrated in Figure~\ref{fig:cohort_intro}, a straightforward approach to dividing cohorts is based solely on explicit features and uses the single derived cohort for different healthcare tasks. Despite being intuitive, applying this approach may not satisfactorily tackle EHR data in practice,
% kp: "approach" or "method" - better be consistent
% wq: revised
as the following two desiderata are not attained: 
%(1) \textbf{Coarse-grained Division.} As shown in Figure~\ref{fig:cohort_intro}(b), Real clinical EHR data reflect the temporally visiting records of patients at the hospital with different diagnosis results, which means the single feature-based cohort may divide one patient into different cohorts at different time points. In realistic diagnosis,  the doctor  usually recalls the health status of similar patients that she/he has treated, or looks up their history visiting records from the hospital system, and then assess and treat the current patient. In other words, the fine-grained cohort requires  considering the patient relevance in time series rather than the single feature. 
(1) \textbf{Fine-grained Cohort Division.} As in Figure~\ref{fig:cohort_intro}(b), real-world EHR data record patients' temporal visits to the hospital for different reasons, thus leading to different diagnoses, medications, etc. Consequently, the coarse-grained cohort division based on a single feature may assign two similar patients to different cohorts at different time points. 
% kp: this sentence above is not finished? why "may assign one patient to different cohorts at different time points" is not good?
In clinical practice, when attending to a patient, a doctor tends to refer to the health conditions and medical histories of similar patients that she/he treated before, to help assess this current patient. In other words, a fine-grained cohort division is highly desired, which takes into account the patient relevance in time dimension rather than based on merely a single feature.
% kp: rewrite (1), pls check above
(2) \textbf{Intra-cohort and Inter-cohort information Exploitation.} As discussed, the relevance of similar patients within a cohort, \emph{i.e.}, intra-cohort information is valuable to medical decisions. Besides, a doctor may also consider the difference between the current patient and other patients with different symptoms, which serves as contrasting knowledge from other cohorts to consolidate the diagnosis.  Therefore, effective EHR representation learning should exploit the intra-cohort and cross-cohort information comprehensively to utilize EHR data to the utmost. However, both the desiderata are not achieved in existing cohort learning  work, such as GRASP~\cite{zhang2021grasp} only considers a simple clustering method to construct cohorts, which may fail to fully exploit the EHR's potential.

% kp: rewrite (2), pls check above; if positioning of the paper on these 2 desiderata is okay and reasonable, then maybe can revise abstract/conclusion accordingly - existing work can do these 2, we can do these 2
% kp: change to desiderata
% 1. fine-grained
% 2. we hope to have both inter- & intra- 
%% wq: revised

% kp: need 1 sentence here, e.g., however, both desiderata are not achieved in existing work... (where are the brief intro of related work such as GRASP?) - this is quite important to show the advantages over baselines, especially GRASP
% wq: added
To fulfill the desiderata above, we propose \method{}, a universal \textbf{CO}hort \textbf{R}epresentation l\textbf{E}arning framework for EHR Data Analysis. As illustrated in Figure~\ref{fig:cohort_intro}(c), \method{} consists of two key steps.
%simulates the aforementioned two aspects by two designs: 
%{\color{red}(I) (ii)}
%%% ooibc?
%that 
% \method{} first introduces a pre-context task  designed to derive cohorts from medical codes and the cohort module, and encode cohort information into the backbones. 
% Unlike GRASP, GRASP utilizes the representations from the backbone model with Gumbel-Max technique to get cohorts, which is not interpretable, 
\textbf{Step 1}, we propose the introduction of a pre-context task that measures the patient relevance to construct fine-grained cohorts. Specifically, we first model the patient-level features by leveraging a patient's sequential visit-level features via the reverse-time attention mechanism. The underlying rationale of this step is that the latter visits in time tend to have higher weights to represent the patient's health status in the implicit semantic space. Thereafter, we measure the Jaccard similarity~\cite{niwattanakul2013using} of any two patients' diagnosis codes as the pseudo label, which serves as the relevance label to model the patient relevance.
% then concatenate their representation to predict the corresponding pseudo label, which can further refine the patient-level representation according to their similarity's label.
\textbf{Step 2}, after modeling the patient relevance backed by the pre-context task, given 
%an EHR embedding of a patient
a patient's embedding
learned from an existing EHR data learning backbone, we distill the intra-cohort information according to her/his similar patients in the same cohort and further incorporate the inter-cohort information for this patient, respectively exploiting the internal and external
% kp: maybe "contrasting" -> "external"?
% wq: revised
information for boosted performance in downstream healthcare tasks. Finally, \method{} exploits the intra-cohort and inter-cohort information as two graphs and encodes them as the augmented representation for prediction. Our contributions are summarized as follows.

 % its cohort module encodes cohort information into the representations of backbones as augmentation from intra-cohort and inter-cohort perspectives.
%%% ooibc: expand this point a bit more

\begin{itemize}
%To address the challenges mentioned above, 
% We propose \method{}, a universal  COhort Representation lEarning framework for EHR data to extract cohorts from medical codes and encode cohort information into representations of backbones.

% \item 
% %\method{} proposes 
% We propose a novel pre-context task to model the temporal EHR data to measure the relevance of patients, thereby constructing the fine-grained cohort.

% \item
% \method{} extracts the information of cohorts from both intra-cohort and inter-cohort perspectives, and comprehensively global cohort information for augmented EHR data representation learning.
% %%% ooibc: english
% %%% fixed the English errors
% % kp: "global"? there is no "local"...

% \item The extensive experimental results show that our proposed \method{} outperforms existing methods by a large margin in readmission tasks on two real-world EHR datasets.

% kp: i'm not quite sure how we should position the contributions - shall we do 2 desiderata? or shall we emphasize the techniques, including pre-context task, 2 graph building, etc? also, how CORE could be applied to different backbones better be mentioned in introduction as well
% kp: i have rewritten the 3 contributions in the 2 desiderata style below - pls check if okay to use?
\item We propose \method{}, a universal cohort representation framework to support fine-grained cohort division and exploit intra-cohort and inter-cohort information simultaneously, which are overlooked in prior studies in EHR analysis.

\item \method{} functions in two steps. Specifically, \method{} first proposes a novel pre-context task to model the temporal EHR data for measuring the relevance of patients, thereby constructing the fine-grained cohorts. \method{} then distills the cohort information from both intra-cohort and inter-cohort perspectives, and finally derives the overall cohort information to facilitate EHR data representation learning.
% kp: "EHR data representation learning" or "EHR representation learning"? better be consistent

\item We evaluate the effectiveness of \method{} on two real-world EHR datasets. The experimental results demonstrate that \method{} on top of backbone models outperforms counterparts by a large margin, validating the efficacy of the comprehensive cohort information learned by \method{}.
% kp: consider adding interpretability part if there are results later

\end{itemize}

The remainder of this paper is organized as follows. We review existing work in Section~\ref{sec:related work}, and present our proposed \method{} framework which achieves fine-grained cohort division and exploits both the intra-cohort and the inter-cohort information in Section~\ref{sec:methodology}. Our extensive experimental results are demonstrated in Section~\ref{sec:experiments}. Finally, we summarize in Section~\ref{sec:conclusion}.

\section{Related work}
\label{sec:related work}

In this section, we review the preliminaries of EHR data representation learning and the corresponding existing studies in this line of research.
% kp: avoid using "pipeline", as CORE is stated as a pipeline in fig1 as well

% kp: i do not feel such preliminaries of EHR data representation learning are necessary - if need to reduce contents later, can shorten this part
EHR data are a systematized collection of patients' information, recording their time-series medical features during the visits to hospitals and are stored in hospitals' systems. EHR data mainly include: (i) structured data such as socio-demographic information, diagnoses, laboratory test results, medications, and procedures, (ii) unstructured data such as medical images (e.g., magnetic resonance imaging data and computerized tomography scans), and doctors' notes.
We categorize the existing machine learning models in EHR analysis as follows and elaborate on several representative studies per category.
% kp: this paragraph is too long - consider divide it
\textbf{Word2Vec-based models}~\cite{mikolov2013distributed}. For instance, medical concept embeddings are learned from healthcare claims data~\cite{choi2016learning}, and Med2Vec~\cite{choi2016multi} utilizes the codes within a visit and a hierarchical structure with a sequential order of visits to learn the representations. 
\textbf{AutoEncoders}.
% kp: "AutoEncoder" or "autoencoder" - need to be consistent
A stack of denoising autoencoders (DAE) is employed by Deep Patient~\cite{miotto2016deep} to learn patient representations from the multi-level clinical descriptors, whereas a sparse autoencoder (SAE) with Gaussian process regression is adopted to deal with dirty clinical data~\cite{lasko2013computational}. \textbf{Convolution Neural Networks}\cite{lecun1998gradient, bouvrie2006notes}. For example, a novel temporal fusion CNN, which extends the connectivity between the layers in the time dimension, is devised to learn temporal features in~\cite{cheng2016risk}, and a hybrid model RCNN, which combines recurrent neural networks (RNN) and convolutional neural networks (CNN), to extract features and preserve the temporal relationship in EEG data in SLEEPNET~\cite{biswal2017sleepnet}.
\textbf{Graph Neural Networks}.
% kp: for consistency, need a ref to GNN
In this category, GRAM~\cite{choi2017gram} proposes a DAG-based attention model to encode the hierarchical information inherent to medical ontologies into EHR data. KAME~\cite{ma2018kame} devises a knowledge attention mechanism that not only learns interpretable representations for the nodes in the knowledge graph but also explores knowledge insights to be used for the model.
% kp: note we do not use simple past tense in cs papers
Besides, CompNet~\cite{wang2019order} employs a graph convolutional model based on medicine combination prediction with a reinforcement learning method to learn the interactions between different medicines. 
\textbf{GAN}~\cite{goodfellow2020generative}. 
% kp: for consistency, need to spell out full name of GAN
Specifically, RCGAN~\cite{esteban2017real} proposes a recurrent GAN and a recurrent conditional GAN to obtain realistic synthetic multi-dimensional time series EHR data. medGAN~\cite{choi2017generating}, on the other hand, 
%generates distributed representations of realistic synthetic patient records via a combination of an autoencoder and generative adversarial networks.
adopts a combination of an autoencoder and generative adversarial networks to achieve a similar goal.
% kp: check if above is ok? "generative adversarial networks" -> "a GAN"?
\textbf{RNN}.
Under this category, Retain~\cite{choi2016retain} uses RNN to calculate the attention of both historical visits and significant clinical variables. Dipole~\cite{ma2017dipole} employs bidirectional recurrent neural networks
% kp: for RNN, CNN, GAN, etc: 1st time appear, use full name + (abbreviation); later all use abbreviation consistently
%to interact with the information of both past visits and future visits. 
to model the temporally dynamic behaviors of EHR data from both directions.
In addition, Dipole introduces three attention mechanisms to measure the relationships of different visits for EHR analysis.
\textbf{Transformer-based models}.
% kp: for consistency, need a ref to GNN
BEHRT~\cite{li2020behrt} treats each diagnosis as words and each visit as a document with the aim of learning the relationships between patients' diagnoses and their previous visits.
%from their visit sequences.
G-BERT~\cite{shang2019pre} constructs ontology embeddings from the diagnosis ontology tree and the medication ontology tree for medical code representation and medication recommendation. Med-BERT~\cite{rasmy2021med}, trained on structured diagnosis data, designs a domain-specific cross-visit pretraining task to capture contextual semantics. ClinicalBERT~\cite{huang2019clinicalbert} learns representations based on high-dimensional and sparse clinical notes and fine-tunes the representations on a readmission task.
% kp: have heavily rewritten the paragraph above - pls check if ok
% kp: need a sentence here, e.g., However, all these studies fail to consider info from similar patients...

\begin{figure*}[t]
  \includegraphics[width=\textwidth]{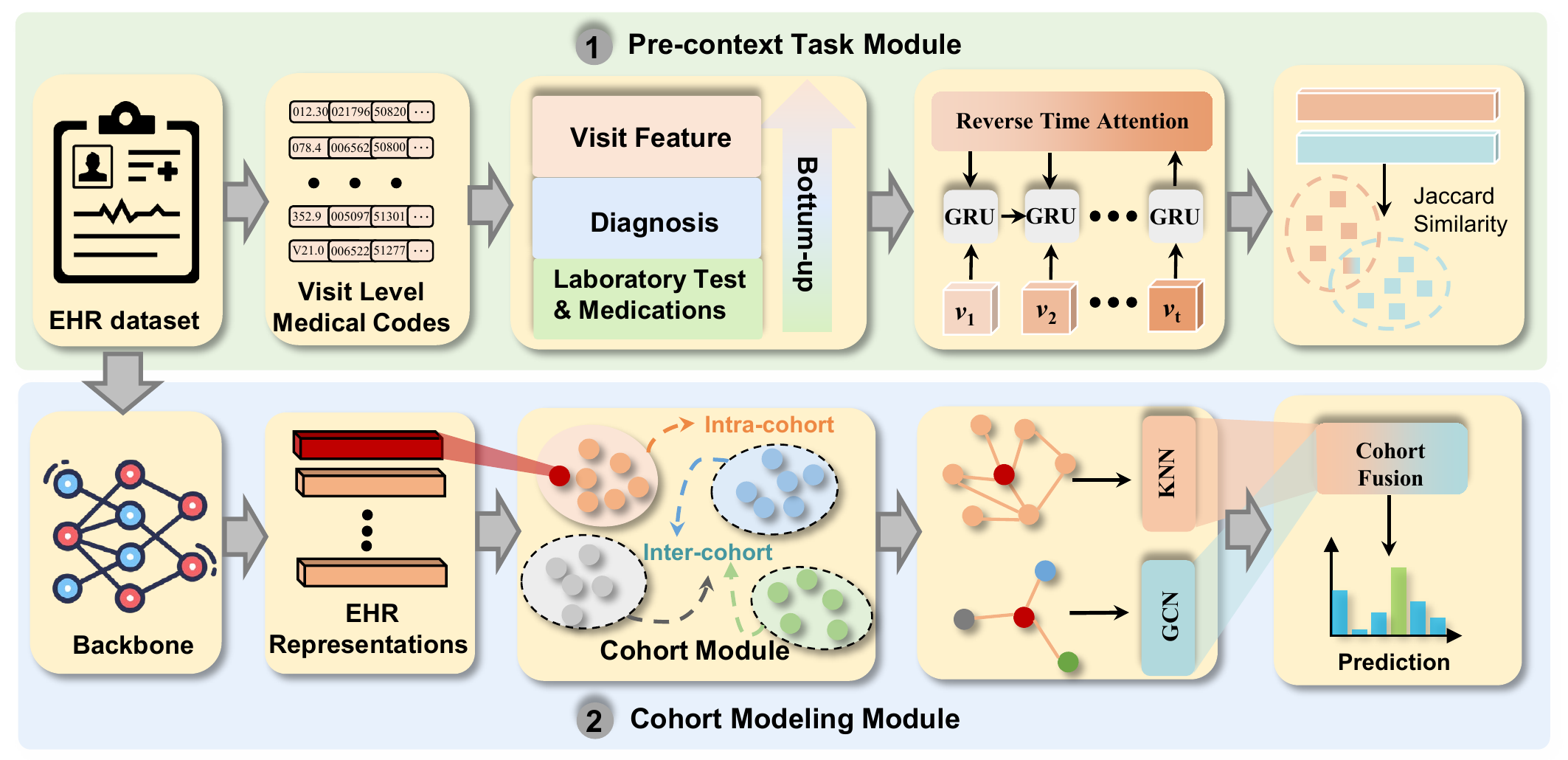}
  \caption{The overview of \method{}. \method{} obtains cohorts from the Pre-context Task Module and then encodes cohort information into the representations from backbones via the Cohort Modeling Module.
  % kp: both modules are proposed by us - should use capital letter per word; "Cohort Module" in the fig should be removed?
  }
  \label{fig:architecture of CORE}
  \Description{The architecture of \method{}}
\end{figure*}

Further, several studies take into account the information/relevance between patients to boost the performance of analytic models,
% kp: generally not use present perfect tense unless very necessary
as similar patients are more likely to exhibit similar behaviors or medical symptoms. For instance, a patient similarity evaluation framework is proposed in~\cite{zhu2016measuring}, preserving the temporal properties in EHR data meanwhile. A convolutional neural network is adopted in~\cite{suo2018deep} to capture locally important information in EHR data. GRASP~\cite{zhang2021grasp} defines the similarity between patients for different clinical tasks via Gumbel-Max technique~\cite{gumbel1954statistical}.
% kp: need to add more details of GRASP (limitations) - as this is the key baseline in comparison
Nonetheless, these methods either do not consider similarities between patients or cannot fully capture the similarity information.
% kp: not clear why they are not interpretable or not general? maybe emphasize what CORE can do, but they cannot do
\section{Methodology}
\label{sec:methodology}
% In this section, we first introduce the notations to better illustrate our method, then we briefly introduce the whole framework. Finally, two modules are introduced in detail.
We describe the proposed \method{} framework in this section by presenting each module and then the training strategy of \method{}.

\subsection{Problem Formulation}
We first introduce some basic notions and terminologies before diving into the details of \method{}.
Given an EHR dataset that has $N_p$ patients $\mathcal{P}=\{p_1, p_2, \cdots, p_{N_p}\}$, each patient $p$ has a temporal visiting sequence $\mathcal{V}^p=\{v^p_1, v^p_2, \cdots, v^p_{N_v}\}$, where $N_v$ is the number of visits. The features of each visit $v$ contain three types of medical codes, namely diagnoses $\mathcal{D}^v$=\{$d^v_1, d^v_2, \cdots, d^v_{N_v}$\}, medications $\mathcal{M}^v$=\{$m^v_1, m^v_2, \cdots, m^v_{N_v}$\} and laboratory tests $\mathcal{L}^v$=\{$l^v_1, l^v_2, \cdots, l^v_{N_v}$\}.
% kp: need to say explicitly that each visit is "v"?
%wq : revised
In this paper, we propose \method{}, a universal cohort representation learning framework for EHR data analysis. As shown in Figure~\ref{fig:architecture of CORE}, we use the following training pipeline $M(\mathcal{P};\Theta_{{P}},\Theta_{\mathcal{C}})$  to demonstrate how \method{} works: 
\begin{equation}
\begin{aligned}
\underbrace{M(\mathcal{P};\Theta_{{P}},\Theta_{{C}})}_{\rm{\method{}}}=  \underbrace{M_P(\mathcal{S}^{\mathcal{P}}|\mathcal{P};\Theta_{{P}})}_{\rm{Pre-context\ Task}} \rightarrow
\underbrace{M_C(\mathcal{C}|\mathcal{S}^{\mathcal{P}};\Theta_{{C}})}_{\rm{Cohort\ Modeling}}
 \label{adv}
\end{aligned}
\end{equation}
where $M_P(\cdot)$
% kp: should be "M_P"?
and  $M_C(\cdot)$ are the models for the pre-context task and cohort modeling with parameters $\Theta_{{P}}$
% kp: should it be $\Theta_{{P}}$ to match the module name?
% wq: revised
and $\Theta_{{C}}$, respectively. The Pre-context Task Module models the patient relevance $\mathcal{S}^{\mathcal{P}}$ based on patients' visit sequences. The Cohort Modeling Module constructs cohorts $\mathcal{C}$, integrating the intra-cohort and inter-cohort information, and then correspondingly encodes the cohort information to attain augmented representations.

% \method{} consists of a \textbf{Pre-context task module}, which extracts cohorts from patients' medical codes, and a \textbf{Cohort module}, which encodes cohorts' information from intra-cohort and inter-cohort perspectives.  

\subsection{Pre-context Task Module}

\begin{figure}[t]
\centering
  \includegraphics[width=0.8\linewidth]{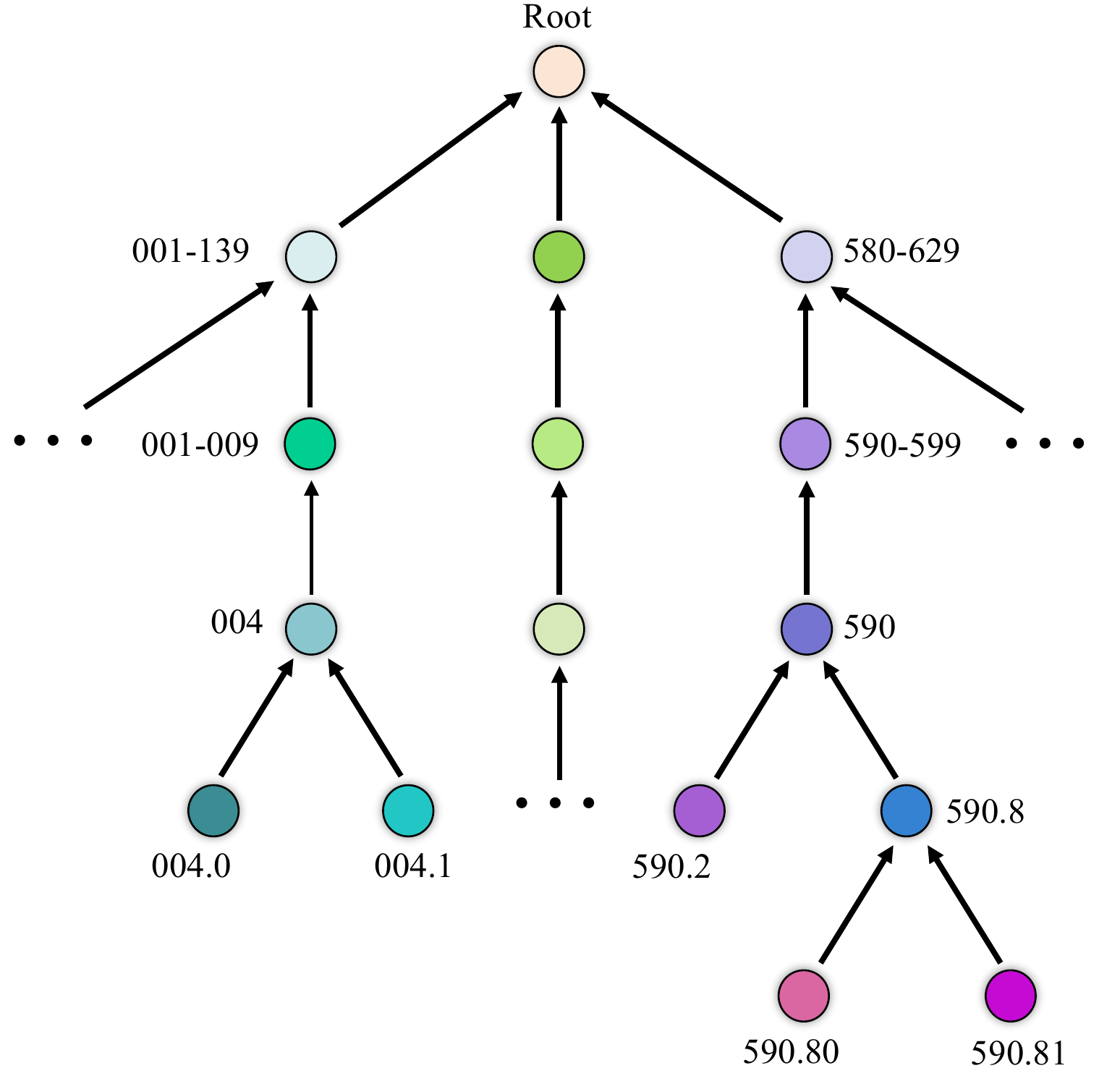}
  \caption{An example of the ICD-9 ontology. The hierarchical architecture of the ICD-9 ontology consists of semantics for codes and the hierarchy for the path from the leaf to the root.}
  \label{fig:ICD-9}
  \Description{Example of ICD-9}
\end{figure}

The Pre-context Task Module aims to obtain cohorts by modeling the patient-level relevance. In brief, \method{} first devises a hierarchical network to process medical codes (diagnosis codes, medication codes, and laboratory test codes) to derive visit-level features of EHR data. \method{} then employs the reverse time attention mechanism to learn the patient-level features based on visit-level features. 
%To make sure the representations are indeed informative enough, 
To facilitate the fine-grained cohort construction, we design a patient-level relevance learning process 
for this  module.
% kp: do we need to explain "pre-context" (with ref) before introducing it 1st time?

\subsubsection{Hierarchical Visit Modeling}
A visit $v$ contains a set of diagnosis codes $\mathcal{D}^{v}$, medication codes $\mathcal{M}^{v}$ and labratory test codes $\mathcal{L}^{v}$. 
% kp: the notations here are too complex, can we just say a visit is "v", and only use "v" as superscript?
Considering the fact that most diagnosis codes follow certain specifications, such as ICD-9~\cite{quan2005coding}, ICD-10~\cite{quan2005coding}, ICD-11~\cite{treede2015classification}, SNOMED~\cite{stearns2001snomed}, etc,
% kp: better give ref
which are medical ontologies, it's of vital significance to model such ontologies. Take ICD-9 ontology as an example, which is an ontology tree as shown in Figure~\ref{fig:ICD-9}. Each leaf in the ICD-9 ontology tree is a standard diagnosis code, \emph{e.g.}, \textit{004.0}. This leaf node along with its ancestors denotes the hierarchical architecture.
%of the leaf. 
For instance, $\textit{Root}\rightarrow\textit{001-139}\rightarrow\textit{001-009}\rightarrow\textit{004}\rightarrow\textit{004.0}$ is the hierarchical architecture of the leaf \textit{004.0}. To distinguish the leaves with similar architectures and derive informative representations of diagnosis codes, we divide the hierarchical architecture of leaves into two parts: semantics and hierarchy.

\begin{itemize}
\item Semantics. Each node in the ontology represents a unique medical term. For example, \textit{580-629} denotes \textit{Diseases Of The Genitourinary System}, \textit{590-599} denotes \textit{Other Diseases Of Urinary System} and \textit{590.81} denotes \textit{Pyelitis or pyelonephritis in diseases classified elsewhere}.

\item Hierarchy. Hierarchy is the path in the diagnosis ontology tree from \textit{Root} to the leaf. For example, the leaves \textit{004.0}, \textit{004.1} and \textit{590.2} have similar hierarchies because their paths are similar.
\end{itemize}

Semantics aims to capture the medical semantics of the diagnosis code, and the hierarchy is used to depict the diagnosis ontology tree. We employ \textit{Node2Vec} and \textit{GloVe} to derive the representation of the diagnosis codes from two perspectives, respectively.

In addition to the hierarchical ontology architecture for diagnosis codes, there exists  another hierarchy among different medical codes. Specifically, in a patient's visit to the hospital, a set of diagnosis codes are assigned to record the presence of diseases, medications are prescribed to cure the disease and laboratory tests are used to help confirm the diagnosis of a disease.
Similar to MiME~\cite{choi2018mime}, we define the hierarchy levels for visit-level representation: \textit{Medications | Laboratory tests} $\rightarrow$ \textit{Diagnoses} $\rightarrow$ \textit{Visits}, as illustrated in Figure \ref{fig:visit-level representation}.

Given a visit $v$ of patient $p$, the modeling of visit-level features is as follows.
\begin{equation}
  F(\mathcal{M}^{v}) = W^Mf^{M^{v}}+b^M
\end{equation}
\begin{equation}
  F(\mathcal{L}^{v}) = W^Lf^{L^{v}}+b^L
\end{equation}
where $F(M^{v})$ and $F(L^{v})$ are the features of medication codes and laboratory codes for visit $v$ of patient $p$, respectively. $M^{v}$ and $L^{v}$ denote the medication codes and laboratory test codes. $f^{M^{v}}$ and $f^{L^{v}}$ correspond to the one-hot representations for medical codes, more concretely, for medication codes and laboratory test codes. $W^M, b^M$, and $W^L,b^L$ are the weights and biases to derive the medication features and laboratory test features, respectively.

\begin{equation}
  F(\mathcal{D}^{v}) = W^Df^{D^{v}}+b^D
\end{equation}
Similarly, the features for diagnosis codes are also obtained by a linear function. 
Different from medication codes and laboratory test codes, $f^{D^{v}}$ for diagnosis codes aggregated below:
% kp: change above to below?
% For modeling diagnosis codes, we also adopt a linear function, but different from medication codes and laboratory test codes, we derive the diagnosis representations $f^{D^{v}}$ via a aggregation below:
% kp: need to explain why diag modeled differently? due to the diag ontology?
\begin{equation}
f^{D^{v}}={ Agg}(H(D^{v}))
\end{equation}
where $Agg$ denotes the aggregation function and in general, we take \textit{mean} as the aggregation function. $H$ denotes the hierarchical modeling for diagnosis codes, which is calculated as:
\begin{equation}
    H(D^{v})=W^H({ Concat}(N(D^{v}), G(D^{v})))+b^H
\end{equation}
\begin{equation}
\label{equ:Node2Vec}
    N(D^{v})={ Node2Vec}(O(D^{v}))
\end{equation}
\begin{equation}
\label{equ:GloVe}
    G(D^{v})={ GloVe}({ Sentence}(D^{v}))
\end{equation}
$Concat$ is the \textit{concatenate} function. $N(D^{v})$ denotes the \textit{Node2Vec} model and $G(D^{v})$ denotes the \textit{GloVe} model. $O(D^{v})$ in Equation \ref{equ:Node2Vec} is the ontology graph of diagnosis ontology. $Sentence(D^{v})$ in Equation \ref{equ:GloVe} is the sentence formed by the semantics of the diagnosis codes in the visit.

Finally, we obtain the visit-level features from the features of diagnosis codes, medication codes, and laboratory test codes.
\begin{equation}
    F(v)_{inter}=W^V_{inter} { Concat} (F(M^{v}), F(L^{v})) + b^V_{inter}
\end{equation}
\begin{equation}
    F(v)=W^V { Concat}(F(v)_{inter}, F(D^{v})) + b^V
\end{equation}
where $F(v)_{inter}$ is the intermediate feature that encodes the medication codes and laboratory test codes and $F(v)$ is the final visit-level features.
% kp: should be "$R(v)$ is the final visit-level representation with diagnosis codes further incorporated"?

\begin{figure}[t]
  \includegraphics[width=\linewidth]{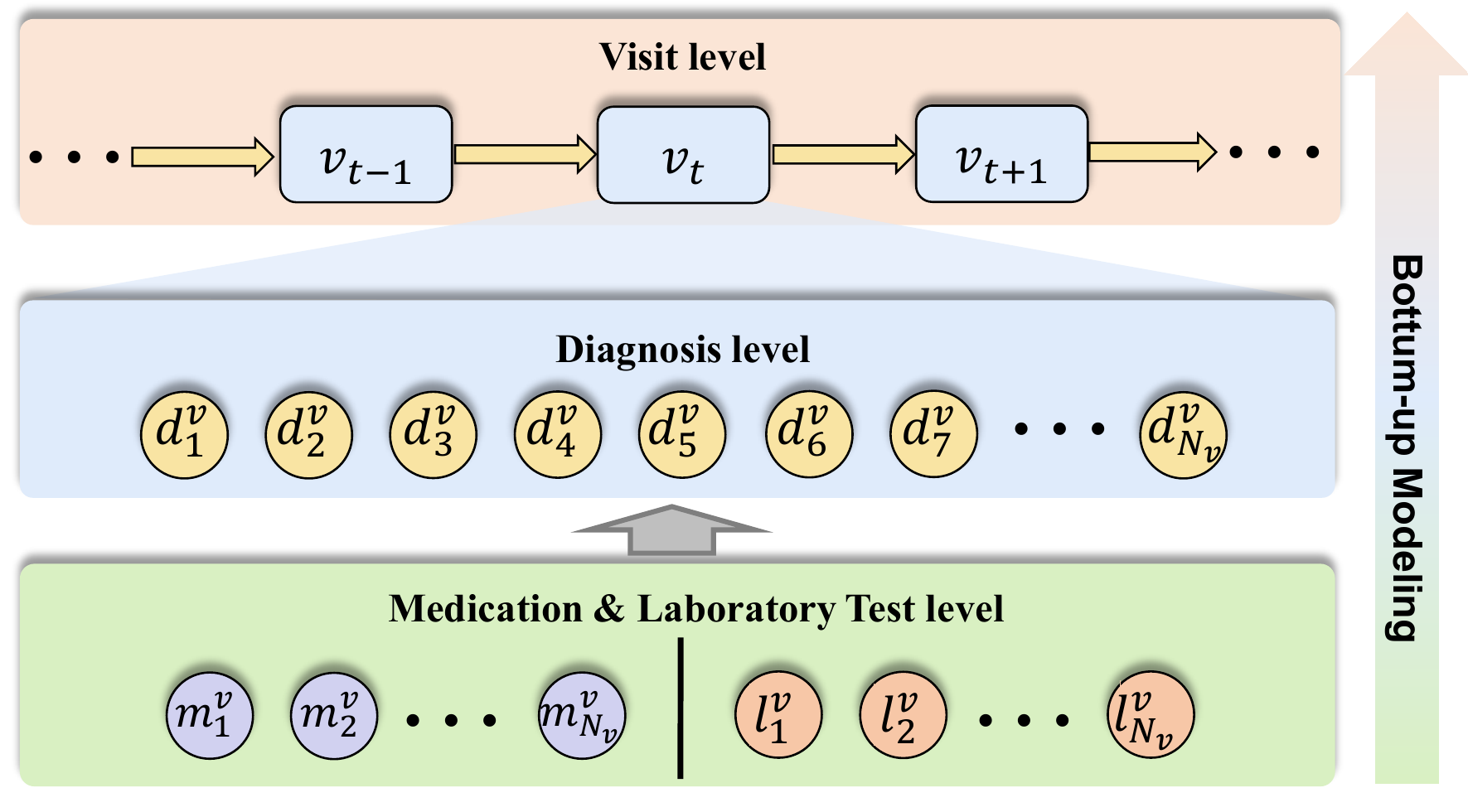}
  \caption{The hierarchy for visit-level features. Medications and laboratory tests are on the same level because they all serve as diagnosis codes. Features of visits are represented by the features from the diagnosis level.}
  \label{fig:visit-level representation}
  \Description{Hierarchy for visit-level features} 
\end{figure}

By doing so, we have attained the visit-level features. In the next step, we shall derive the patient-level features and construct a patient similarity task to make the features informative correspondingly. 
% kp: is patient similarity task semi-supervised or unsupervised? in sec 3.2 - it says unsupervised?

\subsubsection{Reverse time attention for patient-level features}

We need to make sure that the visit-level features learned in \method{} manage to capture the similarity semantics for patients so that we can obtain cohorts from the visit-level features. We devise a patient similarity task to achieve this goal. First, we define the similarity between patients as the Jaccard similarity of their diagnosis codes with all their visits.
% kp: is patient similarity task semi-supervised or unsupervised? in sec 3.2 - it says unsupervised?

\begin{equation}
    sim(p_i,p_j) = Jaccard( \mathcal{D}^{\mathcal{V}^{p_i}}), set(\mathcal{D}^{\mathcal{V}^{p_j}})
\end{equation}
where $Jaccard$ is the Jaccard similarity function, $\mathcal{D}^{\mathcal{V}^{p_i}}$ denotes all the diagnosis codes among all the visits of patient $i$.

Then for each patient, to make the labels balanced and volume appropriate, we select the most similar 5 patients as the positive samples and randomly select 5 negative samples from the remaining patients.
To learn comprehensive patient-level features, we propose to adopt the reverse time attention mechanism suggested by RETAIN~\cite{choi2016retain}, based on the rationale that in clinical practice, doctors will check a patient's medical history for accurate diagnosis, and the closer a historical visit is, 
% kp: "the closer" to what?
the more important the visit is. Therefore, we take the reverse time order of the visits to calculate the attention for each visit and finally obtain the patient-level features.

\begin{equation}
    g^p_{N_v}, g^p_{N_v-1}, \cdots , g^p_1 = GRU(F(v^p_{N_v}), F(v^p_{N_v-1}), ..., F(v^p_1))
\end{equation}
\begin{equation}
    e^p_i = W^{Re}g^p_i+b^{Re}, i= 1,2,3, \cdots ,N_v
\end{equation}
\begin{equation}
    \alpha_{N_v},\alpha_{N_v-1}, \cdots , \alpha_1={ Softmax}(e^p_{N_v}, e^p_{N_v-1}, \cdots , e^p_1) 
\end{equation}
where $GRU$ is used to compute the reverse time features for each visit, $W^{Re}$ and $b^{Re}$ are used to map the reverse time features to a scalar which is then used by the $Softmax$ function to derive the attention.
Subsequently, the patient-level feature is obtained:

\begin{equation}
    F^p = \sum^{N_v}_1 \alpha^p_i F(v^p_i)
\end{equation}

Such patient-level features will be used to predict whether two patients are similar against the labels we defined before. Below is the loss function for the patient similarity task.

\begin{equation}
    \hat{y}_{i,j}= f(R^i,R^j)
\end{equation}
\begin{equation}
    L_{pre} = BCE(y_{i,j} , \hat{y}_{i,j})
\end{equation}
where $\hat{y}_{i,j}$ is the perdiction for similarity between patient $i$ and $j$, $f$ is the classifier. $L_{pre}$ is the loss for pre-context task, $BCE$ is the Binary Cross Entropy loss~\cite{ruby2020binary} function, $y_{i,j}$ is the truth label for patient $i$ and $j$, which we obtain before according to patients' similarity.

While training on the patient similarity task,
% kp: seems there are different expressions of "patient similarity task" before - better be consistent
we obtain cohorts from the patient-level features in the meanwhile which will be used in the downstream task via clustering:

\begin{equation}
    C=Cluster(F^P)
\end{equation}
where $C$ denotes all the learned cohorts, $F^P$ represents the features of patients.

\subsection{Cohort Modeling Module}

Cohorts play a crucial role in EHR analysis in healthcare. However, traditional cohorts are coarse-grained and mainly constrained to capturing intra-cohort information. 

After we obtain the fine-grained cohort information at patient-level  from the pre-context module, 
%%% ooibc: english!
 we set to build the intra-cohort and inter-cohort to maximally utilize the EHR data for healthcare tasks. 
 Note that if visit-level cohorts are desired, we can use the visit-level representations before the reverse time attention mechanism.
 % kp: "we could use the reverse time attention mechanism to obtain the patient representations" -> we already do so?
 % kp: suggest: change "we could use the reverse time attention mechanism to obtain the patient representations before clustering." -> "we could conduct clustering on top of the derived patient representations instead."

%%% ooibc: english
% kp: why choose to do cohort modeling at the visit level? especially that in Eq(15), already learn patient representations based on visit representations, then in Eq(16), choose to cluster visit representations back?

\subsubsection{Individual Representation Learning}
Before we construct the cohort, we first extract the individual EHR data representation as the patient-level embedding. Given a patient $p \in \mathcal{P}$, the initial representation is learned as follows:
% We obtain the representations from the backbones, which could be either visit-level or patient-level representations, and in this paper, we will take visit-level representations to illustrate our method.
% For patient-level representation, we only need to change the visit-level representations before generating cohorts to patient-level representations with the reverse time attention mechanism.

\begin{equation}
  R_{ini}^p = { Backbone}(p)
\end{equation}
where $R_{ini}$ represents the initial representations calculated from the backbones~\cite{choi2016multi,choi2018mime,huang2019clinicalbert} and will be utilized by our \method{}. 

\subsubsection{Intra-cohort Module}
% kp: for consistency, better "Intra-cohort Graph Building/Construction"
% wq: changed

To model the interaction of visits in the same cohort,
% kp: again - why cohort modeling on visits?
we view each cohort as a graph, in which each node represents a visit. We then use the graph aggregation method to update the nodes. In the graph, every pair of nodes is connected. To take advantage of the information of similar patients, we propose to utilize two hyperparameters $\gamma$ and $K$ to obtain dynamic neighbors.
For a node in the graph, we continuously select the nearest neighbors until the number of selected neighbors reaches $K$, 
under the premise that the similarity between the node and the selected neighbor is no lower than $\gamma$, just as the following equation.
% kp: change to: "under the premise that the similarity between the node and the selected neighbor is no lower than $\gamma$:"?

%%% ooibc: not a proper sentence

\begin{equation}
    S^p+=\left\{
    \begin{array}{ll}
        N, N=nearest(C^p - S^p, v) & sim(p, N) > \gamma \\
        & and |S^p| < K \\
        stop & otherwise
    \end{array}
    \right.
\end{equation}
where $S^p$ denotes the selected neighbors for the patient $p$. $N$ denotes the nearest candidate neighbor. $C^p$ is the cohort to which the patient $p$ belongs, excluding $p$ itself.
Subsequently, we employ the $mean$ function as the $Agg$ function to aggregate the representations from selected neighbors to update the representation of $p$.

\begin{equation}
    R^p_{intra}=Agg(S^p, R^p_{ini})
\end{equation}

\subsubsection{Inter-cohort Module}
% kp: for consistency, better "Inter-cohort Graph Building/Construction"

After we obtain the intra-cohort information for each visit, we want to capture the inter-cohort information, which reflects the difference between cohorts. 
% kp: if only centroid part is similar to GRASP - then can remove "Similar to GRASP,"?
We view the centroid of each cohort as a node in the inter-cohort graph. To model the edge relationships, we select the nearest $S$ neighbors for each node and connect them in the graph, with $S$ as a hyperparameter to specify. We then employ the GCN method to derive the representation of each node in the graph.

\begin{equation}
    R_{inter}= GCN(Graph(C, S))
\end{equation}

Finally, we integrate the information from the initial representations by backbones, intra-cohort graph, and inter-cohort graph based on learnable attention as the final representation, which serves as the enhanced EHR data representation for the downstream healthcare tasks. 
% kp: change to: "which will serve for downstream tasks:"

\begin{equation}
    R^p_{final}= R^p_{ini} + att_{intra} * R^p_{intra} + att_{inter} * R^p_{inter}
\end{equation}
\begin{equation}
    att_{intra}, att_{inter}= Softmax(Att_{unit}(R^p_{ini},R^p_{intra},R^p_{inter}))
\end{equation}
\begin{equation}
    Att_{unit}(R^p_{ini},R^p_{intra},R^p_{inter})=MLP(R^p_{ini},R^p_{intra}),MLP(R^p_{ini},R^p_{inter})
\end{equation}
where $R^p_{final}$ is the final representation for prediction, $att_{intra}$ and $att_{inter}$ are the attention for intra-cohort and inter-cohort representations, respectively. We utilize a learnable attention mechanism with softmax function and MLP to learn attention. Finally, the loss function for \method{} is as follows.

\begin{equation}
    L=L_{downstream}+ \lambda_{pre}L_{pre}
\end{equation}
where $L_{downstream}$ is the loss function of the downstream task, more specifically, readmission prediction with the $BCE$ loss function. $\lambda_{pre}$ is the hyperparameter for the loss of the pre-context task, which is set to 0.1 in our experiments. Let $f_{downstream}$ denote the classifier adopted in the downstream task. The overall loss function is expressed as:
\begin{equation}
    L=BCE(f_{downstream}(R^p_{final}))+ \lambda_{pre}L_{pre}
\end{equation}

% kp: "Intra(R^{p_v}_{ini})" and "Inter(R^{p_v}_{ini})" should be added before, i.e., sec3.4.1 should end with Intra() formula, sec3.4.2 should include Inter() formula

% kp: but prediction part is not shown?
\section{Experiments}
\label{sec:experiments}

In this section, we describe the two datasets for evaluation and introduce the backbones and baselines adopted for comparison with \method{}. We then describe
the experimental settings and present the experimental results.

\subsection{Datasets}
\begin{itemize}
\item \texttt{MIMIC-III}~\cite{johnson2016mimic}: 
MIMIC-III is a large public medical database, which consists of 58,976 unique hospital admissions from 38,597 patients in the Beth Israel Deaconess Medical Center between 2001 and 2012. Each admission contains a varying number of ICD-9 diagnosis codes, medication codes, and laboratory test codes. There are 2,083,180 de-identified notes associated with the admissions.

\item \texttt{UCI Diabetes} - Machine Learning Repository~\cite{strack2014impact}:
% kp: name is "UCI Diabetes: Machine Learning Repository"? remove "Machine Learning Repository"?
The dataset is a public medical dataset, which is prepared to analyze factors related to readmission as well as other outcomes pertaining to patients with diabetes. This dataset records 10 years (1999-2008) of clinical care at 130 US hospitals and integrated delivery networks. \texttt{UCI Diabetes} includes over 50 features representing patient and hospital outcomes.
% kp: "patient and hospital outcomes"?
It is comprised of 71,518 patients and 101,767 visits, where each visit contains at most 3 diagnosis codes. The diagnosis codes are ordered by importance, i.e., the first diagnosis code is the primary diagnosis for the patient.

\end{itemize}

The detailed statistics of both datasets are listed in Table~\ref{tab:dataset}. We note that the UCI Diabetes dataset does not contain any medications or laboratory test codes; hence, we drop some backbones which are related to the modeling of the two kinds of medical codes.
% kp: maybe better to be more specific - we not use ClinicalBERT as it involves the modeling of med and lab

\begin{table}
  \caption{Statistics of UCI Diabetes and MIMIC-III.}
  \small
  \label{tab:dataset}
  \begin{tabular}{lcc}
    \toprule
    Dataset & UCI Diabetes & MIMIC-III\\
    \midrule
    \# of patients & 71580 & 38597\\
    \# of visits & 101767 & 53423\\
    Avg. \# of visits per patient & 1.42 & 1.38\\
    \midrule
    \# of unique diagnosis codes & 915 & 6984\\
    \# of unique medication codes & 0 & 4686\\
    \# of unique laboratory test codes & 0 & 726\\
    Avg. \# of diagnosis codes per visit & 2.98 & 12.2 \\
    Avg. \# of medication codes per visit & 0 & 78\\
    Avg. \# of laboratory test codes per visit& 0 & 521\\
  \bottomrule
\end{tabular}
\end{table}

\subsection{Backbones and Baselines}

%Our approach
\method{} targets to capture the essential cohort information and encode it into representations. 
Our pre-context task extracts cohorts from medical codes; therefore, we need to consider the representations from backbones with/without the information of medical codes to validate the effectiveness of \method{}.
% kp: why the causal relationship above holds?
% kp: change to "As \method{} functions as a plug-in to enhancing the representations of diverse backbone, we compare the performance of backbones against that of backbones integrating the cohort information from \method{} to validate the effectiveness of our framework."?
We select the following traditional or state-of-the-art methods as the backbones to calculate the initial representations.

\begin{itemize}
\item Med2Vec~\cite{choi2016multi}: 
%A traditional method, that 
Med2Vec not only models the co-occurrence information of medical codes but also constructs a hierarchical structure with a sequential order of visits to learn the representations. We consider it a traditional method.
\item MiME~\cite{choi2018mime}:
MiME constructs a hierarchical architecture that splits the representations into treatment level, diagnosis level, visit level, and patient level, to render the learned representations interpretable and effective. There are no treatment-level codes in the \texttt{UCI Diabetes} dataset; therefore, we change the operation of interactions between the diagnosis level and the treatment level to diagnosis pair-wise interactions.
\item ClinicalBERT~\cite{huang2019clinicalbert}:  ClinicalBERT is an application of BERT on healthcare, which only takes into account medical notes. In view of the long sequence of medical notes, ClinicalBERT splits the medical notes of a visit into fixed-length slices, meaning that the slices within a visit or a patient share the same label. 
Predictions for patients are computed by binning the predictions on each slice.
% kp: the sentence above is hard to understand

\end{itemize} 

\begin{table}[t]
    \centering
    \small
    \caption{Overall performance on the MIMIC-III dataset.}
    % kp: there is no "qualitative experiments" - so better not mention "Quantitative"
    \label{tab:MIMIC-III dataset}
    \begin{tabular}{l|c|c|c|c|c}
 \toprule[1.5pt]
        \multirow{2}{*}{\textbf{Model}} & \multicolumn{5}{c}{{\texttt{MIMIC-III Dataset}}}\\
        \cline{2-6} & \textbf{AUPRC} & \textbf{Accuracy} & \textbf{Precision} & \textbf{Recall} & \textbf{F1} \\  \midrule[1pt]
        \midrule[1pt]
        ClinicalBERT~\cite{huang2019clinicalbert} & 0.630 & 58.7\% & 0.602 & 0.676 & 0.637 \\ 
        \quad + KNN & 0.628 & 58.6\% & 0.604 & 0.665 & 0.632 \\ 
        \quad + K-Means & 0.629 & 58.4\%  & 0.601 & 0.663 & 0.631 \\ 
        \quad + GRASP & 0.618 & 56.2\% & 0.617 & 0.479 & 0.539 \\ 
      \rowcolor{gray!40} \textbf{\quad + \method{}(Ours)} & \textbf{0.658}  & \textbf{60.6\%} & \textbf{0.645} & \textbf{0.629} & \textbf{0.637} \\ \midrule[1pt]
        Med2Vec~\cite{choi2016multi} & 0.540 & 57.4\% & 0.592 & 0.325 & 0.419 \\ 
        \quad + KNN & 0.553 & 57.1\% & 0.550 & 0.360 & 0.435 \\ 
        \quad + K-Means & 0.543 & 54.9\% & 0.536 & 0.360 & 0.430 \\ 
        \quad + GRASP & 0.536 & 55.1\% & 0.541 & 0.346 & 0.422 \\ 
        \rowcolor{gray!40} \textbf{\quad + \method{}(Ours)} & \textbf{0.605} & \textbf{59.7\%} & \textbf{0.593} & \textbf{0.474} & \textbf{0.527} \\ \midrule[1pt]
        MiME~\cite{choi2018mime} & 0.542 & 57.0\% & 0.574 & 0.355 & 0.439 \\ 
        \quad + KNN & 0.535 & 57.0\% & 0.564 & 0.404 & 0.471 \\ 
        \quad + K-Means & 0.540 & 56.8\% & 0.565 & 0.382 & 0.455 \\ 
        \quad + GRASP & 0.556 & 58.6\% & 0.587 & 0.430 & 0.496 \\ 
        \rowcolor{gray!40} \textbf{\quad + \method{}(Ours)} & \textbf{0.595} & \textbf{60.5\%} & \textbf{0.584} & \textbf{0.579} & \textbf{0.581} \\
        % G-BERT~\cite{shang2019pre} & 0 & 0 & 0 & 0 & 0 \\ 
        % \quad + KNN & 0 & 0 & 0 & 0 & 0 \\ 
        % \quad + K-Means & 0 & 0 & 0 & 0 & 0 \\ 
        % \quad + GRASP & 0 & 0 & 0 & 0 & 0 \\ 
        % \rowcolor{gray!40} \textbf{\quad + \method{}(Ours)} & 0 & 0 & 0 & 0 & 0 \\ 
        \bottomrule[1.5pt]
    \end{tabular}
\end{table}
%Our method aims to extract the cohort information. 
Further, we adopt KNN, K-Means, and GRASP as baselines for comparison with our proposed \method{}.
% kp: we position CORE as a framework, not a method

\begin{itemize}
\item KNN: For each node, KNN selects the nearest $K$ nodes as its candidate neighbors. KNN is employed with the mean function to update the representation of the center node.
\item K-Means: K-Means divides all the nodes into $K$ groups. 
% kp: "group" -> "cluster"?
% kp: better use different hyperparameter notation for KNN and K-Means
The nodes in the same group share the same neighbors. We will randomly select $K$ neighbors to update the representation of the center node.
% kp: "center node" -> "centroid"?
We note that the number of clusters in K-Means is the same as the number of cohorts in \method{} and the number of selected neighbors in K-Means is the same as that in the intra-cohort modeling of the Cohort Modeling Module.
% kp: there is no "Intra-cohort Module"
\item GRASP~\cite{zhang2021grasp}: 
GRASP is designed to utilize the information of similar patients.
%However, GRASP obtains cohorts 
GRASP obtains cohorts from the representations of backbones via the Gumbel-Max technique and employs GCN with an inter-cohort graph to enhance representation learning. But GRASP only considers the relationship between cohorts, which limits the performance.
% GRASP is not interpretable, and is more of a clustering method rather than the cohort study in healthcare. 
% %interpretability is of vital importance in healthcare. 
% Further, GRASP only captures the inter-cohort attention. 
% changshuo: revised, but maybe need to polish

%We additionally model the intra-cohort relationship.
\end{itemize} 

\subsection{Experimental Settings}

\subsubsection{Readmission Prediction}
We investigate the readmission prediction task on both datasets. \texttt{UCI Diabetes} dataset is collected to analyze factors related to readmission as well as other outcomes pertaining to patients with diabetes. 
Given a visit, we need to predict whether the patient will be readmitted to the hospital within 30 days. \texttt{MIMIC-III} is a large EHR dataset without the readmission labels, so we calculate the label for each visit. Similarly, given a visit, the target is to predict whether a patient will be readmitted to the hospital in 30 days.

\subsubsection{Experimental Setup}

We train the Pre-context Task Module and Cohort Module in parallel. For deriving the cohorts, we employ the agglomerative clustering
% kp: should be "agglomerative clustering"?
and adjust the number of cohorts $c = 900, 1100, 1300, 1500$ to explore the influence of the size of cohorts. In the Cohort Modeling Module,
% kp: "Cohort Module" -> "Cohort Modeling Module" throughout the paper
we select $\gamma$ from 0.99 to 0.81 and $K$ from 5 to 50 to control the number of selected neighbors in the intra-cohort graph and select $S$ from 5 to 50 to explore the relationship between different cohorts. In addition, in the Cohort Modeling Module, we use the Adam optimizer with the initial learning rate $lr = 1e-3$. We randomly split each dataset into training, validation, and test sets with a ratio of 8: 1: 1. We train 50 epochs in all datasets.
% kp: "in all"?
For each model on each dataset, 
% kp: "For each task and each model," -> "For each model on each dataset,"?
we report $AUPRC$, $accuracy$, $precision$, $recall$, and $F1$ score to evaluate the performance of the models.
% kp: note that auprc != auroc; when introducing auroc, auprc for 1st time, better add ref; how many repeats?
% kp: shall we introduce the oversampling conducted to render the label balanced? so that reviewers can understand the metrics better

\begin{table}[t]
    \centering
    \small
    \caption{Overall performance on the UCI Diabetes dataset.}
    \label{tab:UCI Diabetes dataset}
    \begin{tabular}{l|c|c|c|c|c}
     \toprule[1.5pt]
        \multirow{2}{*}{\textbf{Model}} & \multicolumn{5}{c}{{\texttt{UCI Diabetes Dataset}}}\\
        \cline{2-6} & \textbf{AUPRC} & \textbf{Accuracy} & \textbf{Precision} & \textbf{Recall} & \textbf{F1} \\ 
        % \textbf{Model} & \textbf{AUPRC} & \textbf{Accuracy} & \textbf{Precision} & \textbf{Recall} & \textbf{F1} \\   
    \midrule[1pt]
    \midrule[1pt]
        Med2Vec~\cite{choi2016multi}& 0.533 & 54.4\% & 0.539 & 0.506 & 0.522 \\ 
        \quad  + KNN & 0.535 & 50.7\% & 0.499 & 0.571 & 0.533 \\ 
        \quad + K-Means & 0.529 & 50.7\% & 0.499 & 0.585 & 0.539 \\ 
        \quad  + GRASP & 0.534 & 53.0\% & 0.524 & 0.492 & 0.507 \\ 
         \rowcolor{gray!40} \textbf{\quad + \method{}(Ours)} & \textbf{0.552} & \textbf{55.3\%} & \textbf{0.539} & \textbf{0.626} & \textbf{0.579} \\ \midrule[1pt]
        MiME~\cite{choi2018mime} & 0.531 & 54.2\% & 0.535 & 0.524 & 0.530 \\ 
       \quad  + KNN & 0.553 & 54.8\% & 0.549 & 0.453 & 0.497 \\ 
        \quad  + K-Means & 0.555 & 54.8\% & 0.548 & 0.461 & 0.501 \\ 
        \quad  + GRASP & 0.533 & 54.7\% & 0.542 & 0.508 & 0.525 \\ 
         \rowcolor{gray!40} \textbf{\quad + \method{}(Ours)} & \textbf{0.555} & \textbf{55.1\%} & \textbf{0.550} & \textbf{0.478} & \textbf{0.511}  \\ \bottomrule[1.5pt]
    \end{tabular}
\end{table}

\begin{figure*}[t]
  \includegraphics[width=\linewidth]{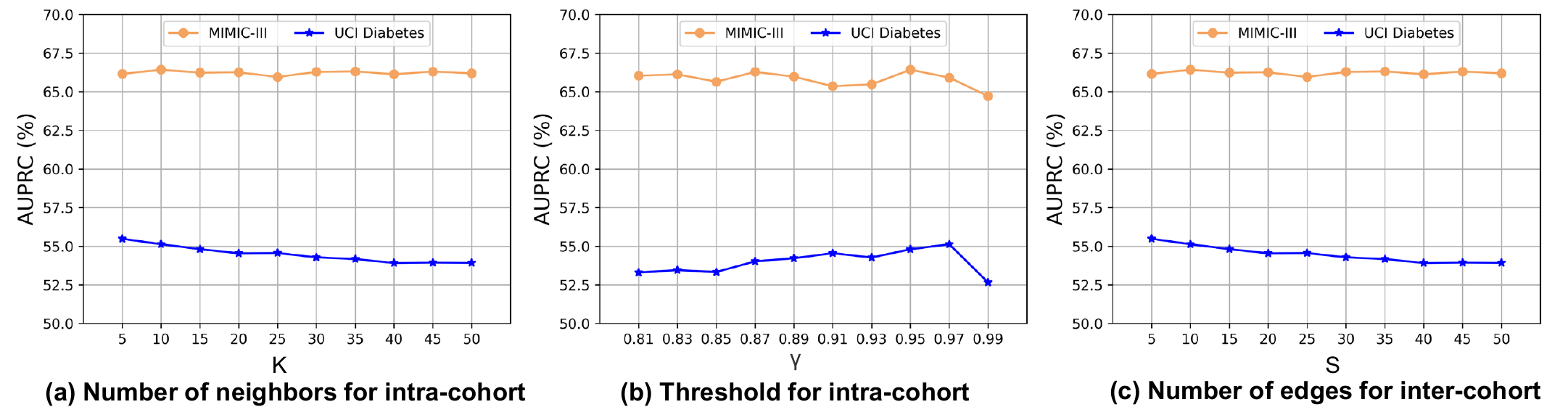}
  \caption{Effects of hyperparameters $K$, $\gamma$ and $S$ on both datasets.}
  \label{fig:gamma and K}
  \Description{Example of ICD-9}
\end{figure*}

\begin{table}[t]
  \caption{The ablation study results of different modules. $M_P$, $M_{intra}$, and $M_{inter}$ represent Pre-context Task Module, Intra-cohort Module, and Inter-cohort Module, respectively.}
  \label{tab:ablation_module_study}
  \centering
 \renewcommand{\arraystretch}{1.2}
 \resizebox{0.48\textwidth}{!}{
{
    \begin{tabular}{c|ccc|c|c|c}
    \toprule[1pt]
    \multirow{2}{*}{\textbf{Model}} & 
    \multicolumn{3}{c|}{{\textbf{Modules}}} & \multicolumn{3}{c}{\texttt{MIMIC-III Dataset}}\\
    \cline{2-7}
    &$M_P$&$M_{intra}$&$M_{inter}$& \textbf{AUPRC} & \textbf{Accuracy} & \textbf{F1} \\
    \midrule[1pt]
    \midrule[1pt]
    \quad -$M_P$ & &\faCheckCircle &\faCheckCircle & 0.635 & 58.2\% & 0.613  \\
    \quad -$M_{intra}$ &\faCheckCircle  & &\faCheckCircle & 0.630 & 58.3\% & 0.630 \\
    \quad -$M_{inter}$ &\faCheckCircle  &\faCheckCircle & & 0.656 & 58.1\% & 0.605 \\
    \midrule[1pt]
    \rowcolor{gray!40} \textbf{$\method{}$} &\faCheckCircle  &\faCheckCircle &\faCheckCircle & \textbf{0.658} & \textbf{60.6\%}& \textbf{0.637} \\
    \bottomrule[1pt]
    \end{tabular}
    }
}

\label{tab:aba}
\end{table}

\subsection{Overall Performance}
\label{sec:overall}
% kp: but ablation study, in-depth analysis, etc should also be experimental results? change subsection title? or merge different experiments as subparts of the experimental results
The overall quantitative results of our framework \method{} and baselines on the test set of the \texttt{MIMIC-III} dataset  and the \texttt{UCI Diabetes} dataset are shown in Table~\ref{tab:MIMIC-III dataset} and Table~\ref{tab:UCI Diabetes dataset}, respectively. From these
tables, we have the following findings: 

\begin{itemize}
\item In general, \method{} achieves the best performance on all the metrics compared to state-of-the-art baselines on both datasets. 
In particular, grounded on the backbone Med2Vec~\cite{choi2018mime}, \method{} outperforms the other baselines in terms of AUPRC by a large margin on the \texttt{MIMIC-III} dataset from \textbf{\underline{5.2\% $\sim$ 6.9\%}}. Notably, the performance improvement on \texttt{MIMIC-III} is more significant than that on the \texttt{UCI Diabetes} (\texttt{UCI Diabetes}: \textbf{\underline{1.7\% $\sim$ 2.3\%}}). The reason for this phenomenon lies in that the patients with the same disease tend to exhibit higher similarity, 
% kp: not quite understand why this explains improvement on MIMIC-III is larger than UCI?
\emph{i.e.}, the pre-extracted patient representations are fairly similar in the implicit semantic space that is sufficiently informative for prediction. Despite this, \method{} demonstrates its superiority over the backbone and other cohort clustering baselines on the \texttt{UCI Diabetes} dataset.
% kp: need a conclusion, e.g., + "which confirms the efficacy of the cohort information unveiled by \methodo{} to facilitate EHR analysis"?

\item  
It is worth noting that the performance of most of the cohort-based approaches yield improved performance over the general EHR data representation learning in the backbones, except for GRASP~\cite{choi2017gram}, which demonstrates that appropriate cohort construction augments the EHR data representation learning, thereby yielding accurate prediction in readmission prediction.
However, compared with backbones, the performance improvement of GRASP is insignificant, even worse than backbones (e.g., Row 1 \emph{vs} Row 4 in Table.2). We conjecture the reason is that GRASP measures the patient similarity via simple clustering, and only utilizes inter-cohort relationships. As a consequence, GRASP fails to model the cohort information in a fine-grained level for accurate readmission prediction.

\item \method{} consistently outperforms all the baselines when using different backbones on both datasets. However, we observe that KNN and K-Means are competitive with \method{} on the \texttt{UCI Diabetes} dataset using the backbone MiME. It's reasonable because we implement the pair-wise diagnosis interactions with MiME on the \texttt{UCI Diabetes} dataset.
% kp: the above sentence is for KNN and K-Means?
This means that the implicit information between diagnosis codes can be incorporated, functioning similarly to our Pre-context Task Module. In this sense, KNN or K-Means resembles our intra-cohort graph module,
% kp: "intra-cohort graph module"? not appear in CORE -> "intra-cohort modeling of Cohort Modeling Module"?
which could account for their achieved competitive performance.
\end{itemize}

\begin{table}[t]
    \centering
    \caption{Cohort case study on the MIMIC-III dataset on the backbone ClinicalBERT, where $MC_A$ is the medical cohorts defined by age. The performance of all involved models are measured by accuracy.}
    \label{tab:Case study}
    \begin{tabular}{l|c|c|c|c|c}
     \toprule[1.5pt]
        \multirow{2}{*}{\textbf{Model}} & \multicolumn{5}{c}{\textbf{Cohort Index}}\\
        \cline{2-6}
        & \textbf{117} & \textbf{277} & \textbf{246} & \textbf{450} & \textbf{812}\\     \midrule[1pt]
    \midrule[1pt]
        $MC_A$ & 77.8\% & 75.6\% & 79.4\% & 75.6\% & 58.1\% \\ 
        GRASP & 69.4\% & 69.8\% & 88.2\% & 66.7\% & 66.1\% \\ 
         \rowcolor{gray!40} \textbf{\method{} (Ours)} & \textbf{85.7\%} & \textbf{79.4\%} & \textbf{91.8\%} & \textbf{81.8\%} & \textbf{69.4\%} \\\bottomrule[1.5pt]
    \end{tabular}
\end{table}
% \begin{figure}[h]
%   \includegraphics[width=\linewidth]{Figure/ICD-9.pdf}
%   \caption{T-SNE of intra-cohort graph for GRASP on MIMIC dataset}
%   \label{fig:intra-cohort graph GRASP}
%   \Description{Example of ICD-9}
% \end{figure}

\subsection{Ablation Study}
% kp: change methodology desc to 3 modules for consistency
We conduct an ablation study on the \texttt{MIMIC-III} dataset to evaluate the importance of each module of \method{} with the experimental results shown in Table~\ref{tab:ablation_module_study}. Comparing \method{} and \method{}(-$M_P$) (Row 1 vs Row 4), the Pre-context Task Module significantly contributes to 2.3\%, 2.4\%  and 2.4\% improvement in terms of AUPRC, Accuracy and F1, respectively. The Intra-cohort Module encodes the information of similar patients, while the Inter-cohort Module captures the difference among cohorts. The results of Row 2 and Row 3 in Table~\ref{tab:ablation_module_study} severally show the performance improvement brought by the Intra-cohort Module and the Inter-cohort Module. Particularly, the huge performance drop caused by removing the Intra-cohort Module confirms the necessity of unveiling the implicit information in similar patients. 
In a nutshell, these results validate the importance of different modules in \method{} for accurate EHR analysis.

\subsection{Sensitivity Study}
% kp: suggest change this subsection to "Sensitivity Study" containing 1st, 3rd; while move Traditional Cohort vs CORE, and Case Study to a new subsection "Case Study" - will involve some changes in each experiment name

% We further probe into several vital perspectives of \method{} in the following analysis.

\noindent\textbf{Effects of Hyperparameters $K$, $\gamma$ and $S$.} We investigate the impact of several crucial hyperparameters on the performance of \method{}: the number of neighbors considered in Intra-cohort Module $K$, the similarity threshold in Intra-cohort Module $\gamma$ and the number of neighbors modeled in Inter-cohort Module $S$. The AUPRC results of varying these hyperparameters on both datasets are shown in Figure~\ref{fig:gamma and K}. From this figure, we observe an interesting phenomenon that the performance of \method{} fluctuates on the \texttt{MIMIC-III} dataset when setting different hyper-parameters. 
On the contrary, the three figures suggest that the optimal choice of $K$, $\gamma$ and $S$ are around 5, 0.97 and 5, respectively, either increasing or decreasing these values will result in a performance decay.
% kp: this is for UCI dataset? then what is the conclusion of the sensitivity study on the MIMIC-III dataset?
% As described in section~\ref{sec:overall}, Table~\ref{tab:UCI Diabetes dataset} has demonstrates the cohort-based representation learning that performs the readmission task better on complex EHR dataset than the dataset with firely similar data. Therefore, 

 % near keeps same compared with backbones. This is reasonable because traditional cohorts are coarse-grained and nearly cannot capture useful information. It goes without saying that fine-grained cohorts are of vital importance.

\vspace{2mm}
\noindent\textbf{Effects of Cohort Number.} In this experiment, we evaluate the impact of cohort numbers. 
%with different backbones 
% kp: "with different backbones" are not emphasized before as well
on both datasets. The corresponding experimental results are illustrated in Figure~\ref{fig:cohort number}. It is clearly shown that the performance of all evaluated models first increases and then decreases as the number of cohorts increases. The reason for such phenomena is related to the target of \method{}, i.e., extracting the information of cohorts to integrate with backbones with boosted performance. If the number of cohorts is smaller, it is harder for the inter-cohort graph to model the difference between cohorts, and the size of cohorts tends to be larger and it's more likely to aggregate dissimilar patients, which leads to noisy information when utilizing the cohorts. On the other hand, with the number of cohorts increasing, fewer patients will be assigned to a cohort, which may result in inadequate cohort information.
% kp: actually "inadequate intra-conhort information"? what about inter-cohort information?
Therefore, the number of cohorts exerts a vitally important influence on the readmission prediction performance on both datasets. 
%We should adjust the number of cohorts to find the best one for different backbones and datasets.
%%% ooibc: it does not add anything
% kp: this sentence above is not finished, add + e.g., what settings we should & have adopted in the experiments (per dataset)
%% ooibc:  expand and explain properly
%%% need more insights
\begin{figure}[t]
  \includegraphics[width=0.9\linewidth]{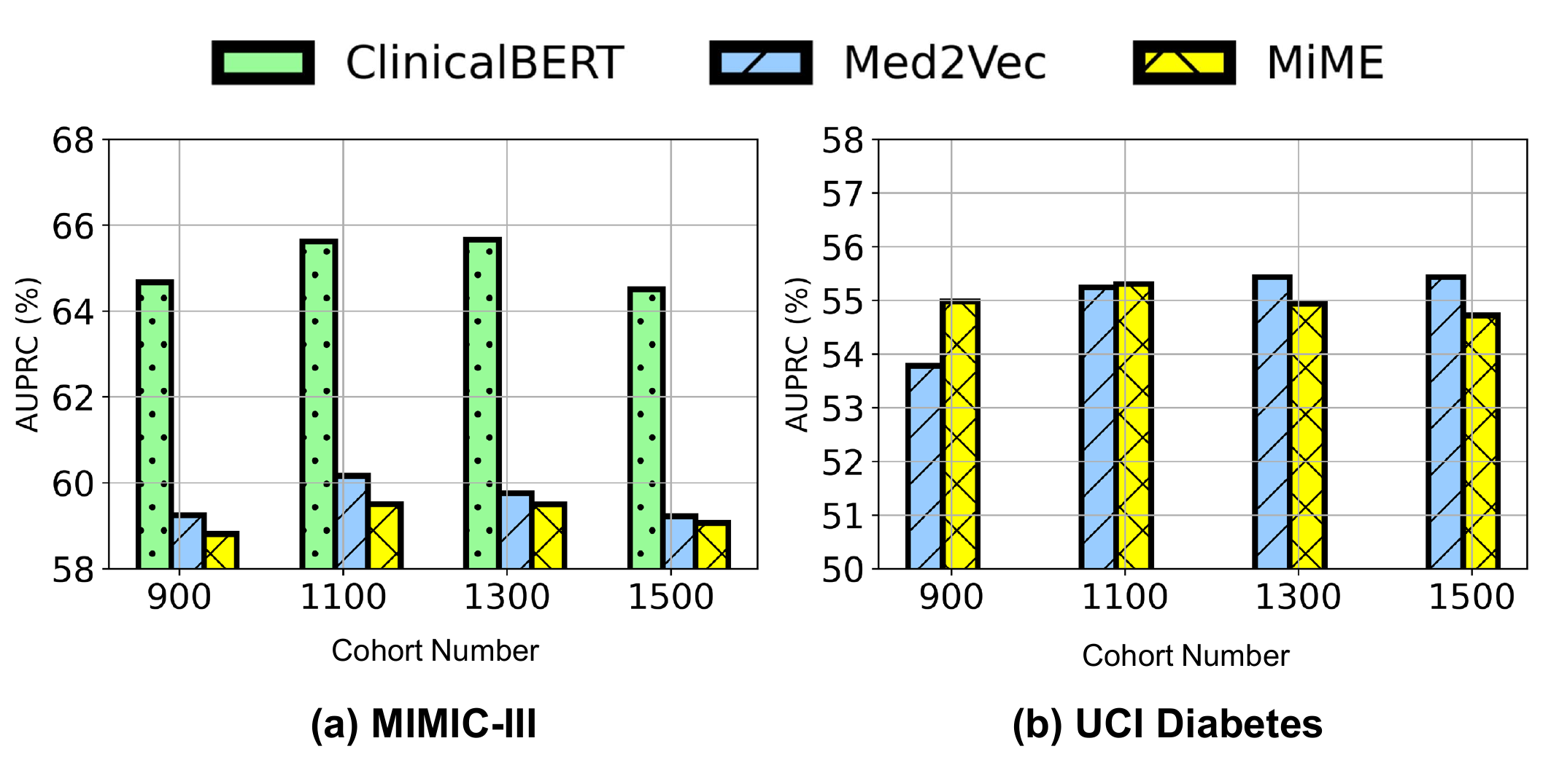}
  % \vspace{-4mm}
  \caption{Effects of the cohort numbers on both datasets.
  }
  \label{fig:cohort number}
\end{figure}
\subsection{Case Study}
\noindent\textbf{Traditional Cohorts \emph{vs} \method{}.} 
To probe into the superiority of \method{} over the traditional medical cohort construction, we further conduct experiments for cohort comparison. We follow the commonly employed cohort division method to extract the medical cohort. Concretely, in our experiment, we choose gender (G), age (A) and both of them (G\&A) as three cohort construction criteria. The comparison results between these medical cohort construction methods and \method{} are displayed in ~Table~\ref{tab:Comparision with traditional cohorts}. As shown, traditional medical cohorts tend to exhibit similar
performance to that of the backbones, some of which are even less accurate than the backbones. This is mainly because the medical cohort construction merely relies on a single feature, resulting in a coarse-grained cohort that is effective enough for cohort pattern mining. Our framework \method{}, on the contrary, leverages a patient’s
sequential visit-level representations from both the intra-cohort and the inter-cohort perspectives at a fine-grained level, thereby boosting the prediction accuracy.

\vspace{2mm}
\noindent\textbf{Cohort Case Study.}
To systematically evaluate the benefits of our proposed strategy of cohort construction, we conduct a cohort case study that chooses five random cohorts with  267 patients from the \texttt{MIMIC-III} dataset. Specifically, we compare the medical cohorts that are split by age, the semantic clustering-based method GRASP~\cite{biswal2017sleepnet} and \method{}. The results are illustrated in Table~\ref{tab:Case study}.
%as contrasting methods. 
%As illustrated, 
We observe that \method{} achieves the highest accuracy in all the selected cohorts.
GRASP yields worse performance than $MC_A$ in 3 out of 5 cohorts because GRASP derives cohorts from the representations of backbones via a clustering technique. 
% kp: "physical meaning"? is it "medical meaning" or "clinical meaning"? Also, as the interpretability of CORE is not evaluated, maybe not good to say GRASP is not interpretable
$MC_A$ obtains cohorts according to age, which is a simple but meaningful strategy since age is a significant predisposing factor for different kinds of diseases. However, $MC_A$ lacks other medical information about the cohorts, which may limit performance.
Summing up, \method{} not only derives a fine-grained cohort construction based on the prior knowledge of medical and diagnosis code, but also exploits the intra-cohort and inter-cohort information, which yield an augmented EHR data representation learning for facilitating healthcare. 
\begin{table}[t]
    \centering
    \small
    \caption{Comparison between traditional medical cohorts and \method{} on the backbone ClinicalBERT. $MC_G$, $MC_A$ and $MC_{G\&A}$ denote the medical cohorts defined by gender (``Male'', "Female", or both genders), age (0-100; the interval is 10, 10 cohorts) and both gender and age (20 cohorts), respectively.}
    \label{tab:Comparision with traditional cohorts}
    \resizebox{0.43\textwidth}{!}{
    \begin{tabular}{l|c|c|c|c|c}
     \toprule[1.5pt]
        \multirow{2}{*}{\textbf{Model}} & \multicolumn{5}{c}{{\texttt{MIMIC-III Dataset}}}\\
        \cline{2-6}
         & \textbf{AUPRC} & \textbf{Accuracy} & \textbf{Precision} & \textbf{Recall} & \textbf{F1} \\  
    \midrule[1pt]
    \midrule[1pt]
    ClinicalBERT & 0.630 & 58.7\% & 0.602 & 0.676 & 0.637 \\ 
        \quad + $MC_G$ & 0.629 & 58.5\% & 0.599 & 0.678 & 0.636 \\ 
        \quad + $MC_A$ &  0.631 & 58.3\% & 0.598 & 0.676 & 0.635 \\ 
        \quad + $MC_{G\&A}$ & 0.626 & 58.4\% & 0.599 & 0.677 & 0.636 \\ 
         \rowcolor{gray!40} \textbf{\quad + \method{} (Ours)} & \textbf{0.658} & \textbf{60.6\%} & \textbf{0.645} & \textbf{0.629} & \textbf{0.637} \\\midrule[1pt]
          Med2Vec& 0.533 & 54.4\% & 0.539 & 0.506 & 0.522 \\ 
        \quad  + $MC_G$ & 0.536 & 54.2\% & 0.529 & 0.530 & 0.523 \\ 
        \quad + $MC_A$ & 0.538 & 53.2\% & 0.519 & 0.598 & 0.536 \\ 
        \quad  + $MC_{G\&A}$ & 0.535 & 52.8\% & 0.517 & 0.564 & 0.531 \\ 
         \rowcolor{gray!40} \textbf{\quad + \method{} (Ours)} & \textbf{0.552} & \textbf{55.3\%} & \textbf{0.539} & \textbf{0.626} & \textbf{0.579} \\ 
         \bottomrule[1.5pt]
    \end{tabular}}
\end{table}

% \begin{figure}[h]
%   \includegraphics[width=\linewidth]{Figure/ICD-9.pdf}
%   \caption{Cohort Size Graph, number of cluster}
%   \label{fig:cohort size}
%   \Description{Example of ICD-9}
% \end{figure}

% \begin{figure}[h]
%   \includegraphics[width=\linewidth]{Figure/ICD-9.pdf}
%   \caption{$\beta$ graph, number of labels in pre-context task}
%   \label{fig:beta}
%   \Description{Example of ICD-9}
% \end{figure}

\section{Conclusions}
\label{sec:conclusion}

Cohort information is essential in EHR analysis in boosting analytic performance. However, two key desiderata, namely fine-grained cohort division and intra-cohort and inter-cohort information exploitation, are not well fulfilled in prior studies.
%In this paper, to fully utilize the information of cohorts, which is useful in healthcare for cohort studies,
To bridge this gap, we propose \method{}, a universal cohort-based representation enhancement framework for EHR data representation learning. \method{} derives cohorts from a Pre-context Task Module and then encodes the information of cohorts  from intra-cohort and inter-cohort perspectives in a Cohort Module. Extensive experimental results show that \method{} outperforms baselines on top of multiple backbones, and provide valuable medical insights into EHR analysis.
% kp: may be not able to say "state-of-the-art" here
% The learned cohorts from \method{} are highly interpretable based on medical codes, which provide valuable medical insights into EHR analysis.
% kp: do not say we are interpretable, and GRASP is not interpretable
%. Hope our study could motivate more researchers to focus on interpretable methods for healthcare.

\newpage
\bibliographystyle{ACM-Reference-Format}
\bibliography{sample-base}

% \appendix









\end{document}